\documentclass[sigconf]{acmart}

\renewcommand\footnotetextcopyrightpermission[1]{} % removes footnote with conference information in first column
\DeclareUnicodeCharacter{200A}{!!!FIX ME!!!}

\AtBeginDocument{%
  \providecommand\BibTeX{{%
    \normalfont B\kern-0.5em{\scshape i\kern-0.25em b}\kern-0.8em\TeX}}}

\setcopyright{none} 

\begin{document}

\title{Combat AI With AI: Counteract Machine-Generated
Fake Restaurant Reviews on Social Media}


\author{Alessandro Gambetti}
\affiliation{%
  \institution{Nova School of Business and Economics}
 % \streetaddress{1 Th{\o}rv{\"a}ld Circle}
  \city{Carcavelos}
  \country{Portugal}}
\email{gambetti.alessandro@novasbe.pt}

\author{Qiwei Han}
\affiliation{%
  \institution{Nova School of Business and Economics}
 % \streetaddress{1 Th{\o}rv{\"a}ld Circle}
  \city{Carcavelos}
  \country{Portugal}}
\email{qiwei.han@novasbe.pt}


\begin{abstract}
Recent advances in generative models such as GPT may be used to fabricate indistinguishable fake customer reviews at a much lower cost, thus posing challenges for social media platforms to detect these machine-generated fake reviews. We propose to leverage the high-quality elite restaurant reviews verified by Yelp to generate fake reviews from the OpenAI GPT review creator and ultimately fine-tune a GPT output detector to predict fake reviews that significantly outperforms existing solutions. We further apply the model to predict non-elite reviews and identify the patterns across several dimensions, such as review, user and restaurant characteristics, and writing style. We show that social media platforms are continuously challenged by machine-generated fake reviews, although they may implement detection systems to filter out suspicious reviews. 

\end{abstract}


%\begin{CCSXML}
%<ccs2012>
%   <concept>
%       <concept_id>10010147.10010178.10010179.10010182</concept_id>
%       <concept_desc>Computing methodologies~Natural language generation</concept_desc>
%       <concept_significance>300</concept_significance>
%    </concept>
%   <concept>
%       <concept_id>10002951.10003227.10003241.10003244</concept_id>
%       <concept_desc>Information systems~Data analytics</concept_desc>
%       <concept_significance>300</concept_significance>
%       </concept>
%\end{CCSXML}

%\ccsdesc[500]{Information systems~Data analytics}
%\ccsdesc[500]{Computing methodologies~Natural language generation}




\maketitle
\pagestyle{plain}

\section{Introduction}
Online reviews have served as valuable signals about product quality that may bridge information asymmetries between customers and sellers in online marketplaces, which in turn influence customer purchases \cite{Duan2008,Hu2008, Ott2012,Vana2021}. Fake reviews can be defined as opinion-based disinformation, which is fabricated and propagated by spammers with the ambition of misleading and deceiving customers \cite{paul2021}. With the prevalence of review systems embedded in social media, such as Tripadvisor, Yelp, and Facebook, fake reviews are also witnessed to proliferate on these platforms, aiming to mislead customers by pretending to be authentic, in order to achieve unjustly competitive gains for certain businesses given the low cost and high monetary gain \cite{Jindal2008,ott-etal-2011-finding,Ott2012}. The COVID-19 pandemic may further exacerbate the issue because many less-experienced customers are forced to make more online purchases and tend to rely on reviews more heavily \cite{McCluskey2022}. According to a report from World Economic Forum, fake reviews' economic impact on global online spending has reached \$152 billion in recent years \cite{Marciano2021}.

Given that the fundamental value of reviews is rooted in their authenticity that reflects customers' truthful experience, fake reviews would not only harm customers but also severely threaten to erode their trust in online reviews and damage the reputation of social media \cite{Ma2014,he2022}. Typically, \cite{he2022} showed that there exists a market of fake reviews, and sellers choose to rely on paid review farms for content creation. To this end, major platforms propose countermeasures to fight review fraud with both manual analyses by the content moderation team and automated systems. For example, since 2019, Tripadvisor has started to publish the transparency report outlining its effort to keep fake reviews off the site. The most recent report in 2021 revealed that over 2 million (3.6\%) of reviews were determined to be fraudulent \cite{tripadvisor2021}. In a similar vein, Yelp implemented an automated recommendation software to filter off 4.3 million suspicious reviews out of 19.6 million reviews and only displayed the most reliable and helpful reviews on the main business pages \cite{McCluskey2022}. 
However, social media still face significant challenges to counteract machine-generated fake reviews effectively due to advances in generative models such as GPT models \cite{openai2023gpt4}. On the one hand, fake reviews fabricated by AI systems trained on real reviews have become essentially indistinguishable. One study showed that machine-generated fake reviews could evade human detection and even receive a higher score of perceived usefulness compared to reviews written by humans \cite{Yao2017}. On the other hand, the cost of machine-generated reviews is considerably lower than buying from sellers of human-crafted fake reviews. 

%{\color{red} Taking into account the cost of fabricating and posting machine-generated fake reviews is of primary importance, because \cite{Ott2012} demonstrated that review platforms %such as TripAdvisor and Yelp with low cost and high monetary gain for posting fake generated content experience higher rates of fake reviews. In light of this, OpenAI currently provides tools such as ChatGPT and their publicly-available API to generate fake content at a low cost, which, in conjunction with wider accessibility, raises concerns about their misuse intended for spamming activities.}

Overall, there is a large body of literature aimed at detecting, explaining, and analyzing how user-generated fake reviews impact social media platforms \cite{wu2020literaturereview,paul2021}. However, no effort has been directed toward explaining, under different heterogeneous metrics, how fake AI-generated reviews have been affecting such platforms thus far, leaving a critical research gap in the literature.
In this paper, we aim to detect and identify the patterns in machine-generated fake restaurant reviews fabricated from high-quality elite reviews on Yelp. As we leverage elite reviews that are verified by Yelp to generate fake reviews using OpenAI's GPT-3 model, our survey results show that humans cannot distinguish AI-generated fake reviews effectively. Furthermore, we implement several fake review detection models and find that the fine-tuned GPT-3 model on our proposed dataset achieves the best performance. Lastly, instead of using filtered reviews as the proxy for fake reviews, we choose to evaluate non-filtered reviews more rigorously, by examining potential fake reviews that are not identified by the Yelp system across several dimensions, including review characteristics, user characteristics, restaurant characteristics, and writing style.
% say something about the findings 
%{\color{red} By separating predicted AI-generated fake reviews from the predicted authentic ones, we performed ANOVA simulations for each attribute, showing that fake reviews score a higher average rating, are posted by users with less established reputations, and are simpler to understand. Very importantly, by leveraging restaurants' official foot-traffic mobility data, we describe how fake reviews are more likely to be associated with restaurants that receive fewer customer visits, which are a robust proxy for restaurant demand. To the best of our knowledge, no study has implemented a similar strategy to pragmatically correlate the impact of fake reviews on customer visits thus far. } % REPHRASE IT 
% REPHRASED

Through ANOVA, we conducted a study that distinguished between AI-generated fake reviews and authentic reviews across each dimension. Our analysis revealed that the average rating of fake AI-generated reviews is higher, and they are typically posted by users with less established reputations. Furthermore, fake reviews are comparatively easier to comprehend than authentic ones. Importantly, we also utilized official foot-traffic mobility data to demonstrate that fake AI-generated reviews are more commonly associated with restaurants that receive fewer customer visits, which is a reliable indicator of restaurant demand. To the best of our knowledge, no prior study effectively established a correlation between fake reviews and customer visits.

%{\color{red} This paper is structured as follows. Section \ref{sec:literature_review} reviews the related literature on the impact and detection of fake reviews on social media platforms. Section \ref{sec:methodology} describes the supervised learning methodologies implemented to detect and analyze fake reviews across several characteristics. Section \ref{sec:results} presents the results. Section \ref{sec:discussion} discusses the results linking back to the current literature, highlighting the limitations as well. Finally, Section \ref{sec:conclusion} concludes the paper.}

%REPHRASED
This paper is organized as follows. Section \ref{sec:literature_review} presents a comprehensive review of the relevant literature regarding the influence and identification of fake reviews on social media platforms. Section \ref{sec:methodology} outlines the supervised learning approaches used to detect and analyze fake reviews across multiple attributes. In Section \ref{sec:results}, the findings are presented, followed by a discussion in Section \ref{sec:discussion}, which links back to the existing literature and highlights any limitations encountered. Lastly, Section \ref{sec:conclusion} concludes the paper.

\section{Literature Review} \label{sec:literature_review}

\subsection{Impact of Fake Reviews in Online Markets}
Different economic agents including retailers and platforms are known for manipulating online reviews \cite{hu2011manipulation, gossling2018manager, lee2018sentiment}. For example, motivated by financial incentives, online merchants are inclined to distribute fake positive reviews for their own products or fake negative reviews against competitors' products \cite{paul2021, crawford2015, wang2018gslda}. Also, online platforms have the propensity to circulate fake reviews to augment website traffic to promote customer engagement \cite{lee2018sentiment}. Remarkably, individual users might also post fake content for reward-seeking purposes \cite{wang2018effects, anderson2014reviews, thakur2018motivates}.
Overall, fake reviews weaken informativeness and information quality \cite{zhang2017welfare}, damaging review credibility and helpfulness \cite{zhang2017welfare, song2017information, zhao2013modeling, sudhakaran2016framework, wan2014reliability, agnihotri2016online}, which are the main factors new consumers take into account when browsing reviews before making purchase decisions. Additionally, extant research has demonstrated that the proliferation of fake reviews increases consumer uncertainty \cite{hunt2015gaming, zhao2013modeling}, inducing customer distrust towards online reviews \cite{filieri2015travelers, zhuang2018manufactured, deandrea2018people}, undermining consumers' purchase intentions \cite{munzel2016assisting, zhuang2018manufactured, fogel2017intentions, xu2020effects}. Specifically, in the context of Yelp, \cite{fogel2017intentions} executed an interdisciplinary study leveraging both qualitative and quantitative research methods such as surveys and linear models, respectively, showing that, from the output of univariate models only, consumers' exact knowledge of review fraud is statistically significantly connected to increased intentions to use Yelp as a tool before making purchase decisions. % here univariate is important 

\subsection{Fake Reviews Characteristics and Detection}
Recent literature examined how fake reviews differed from legit ones across several characteristics such as writing style (e.g. readability and sentiment), ratings, restaurant characteristics, and user behavior. For each characteristic, we succinctly discuss its relevant related work.

\subsubsection{\textbf{Writing Style.}}
\cite{korfiatis2008readbability} posited that reviews' readability serves as a proxy for their helpfulness, as consumers must first read and then comprehend the text to assess their usefulness. Empirical research has further demonstrated that the likelihood of a review being deemed helpful increases when it is presented in an easily comprehensible manner \cite{cao2011exploring}.
Hence, several studies theorized that fraudsters might deliberately disseminate simple fake content to quickly catch readers' attention \cite{li2013helpfulness, agnihotri2016online, vasquez2014discourse}, conceptualizing that fake reviews were easier to comprehend. Empirically, leveraging readability metrics such as the \textit{Automated Readability Index} \cite{senter1967}, \cite{harris2012detecting} found that fake deceptive reviews exhibited less writing complexity as compared to truthful ones. However, no unanimous academic consensus has been established on this finding, because other studies employing comparable methodologies showed the opposite result \cite{Yoo2009, Banerjee2014}. 

Also, textual review sentiment has been investigated for its effectiveness and helpfulness \cite{schindler2012perceived, tang2014neutral}. As for fake reviews, consumers realized that more polarized sentiment tones could be surrogates for suspicious user-generated content \cite{liljander2015young}. For example, leveraging statistical tools such as standard t-tests and generalized linear models, prior research discovered that fake reviews were richer in positive cues as compared to authentic ones \cite{Banerjee2014, Yoo2009}.  
Additionally, \cite{li2011learning} adopted machine learning techniques to identify review spam, concluding that mixed or neutral sentiments were associated with truthful reviews. Also, employing ranking classification models, \cite{liu2018unified} described how spammers are not capable of expressing true sentiment when writing fake reviews, leading to more polarized opinions in the end.

\subsubsection{\textbf{Ratings and Restaurant Characteristics.}}
Extreme sentiment polarity was also detected when considering review ratings \cite{luca2016}, which are a robust representative of sentiment as well. In particular, extant literature affirmed that positive fake reviews were more prevalent than negative ones \cite{lappas2016impact, zhang2019s}. For example, \cite{lappas2016impact} found that 56\% of fake reviews were positive (4-5 stars) and that 29\% were negative (1-2 stars). One hypothesis that may ex-post explain the prevalence of positive fake content could be that a one-star increase in the Yelp restaurant average rating is associated with a 5-9\% revenue growth \cite{luca2016reviews}. 
As far as restaurant characteristics, \cite{luca2016} examined how fake restaurant reviews were present on Yelp. They found that about 16\% of the reviews were filtered out as fake or suspicious, and that restaurants with fewer associated reviews were more likely to submit positive fake reviews to enhance their reputation. 
\cite{luca2016} also segmented restaurants into chain (e.g. McDonald's, Burger King, Subway, etc.) and non-chain, finding the former ones less likely to display positive fake content, because their revenue is not significantly affected by their rating \cite{luca2016reviews}, and because they may incur high reputation costs if caught \cite{mayzlin2014promotional}.

\subsubsection{\textbf{User Behavior.}}
Fake reviews can also be identified by user behavior, i.e. spammers' characteristics. For example, \cite{sandulescu2015detecting} describe the concept of \textit{singleton reviews}, which is the phenomenon of users posting only one fake review. Because of that one-to-one relationship, spotting and tracking activities of singleton review spammers is challenging \cite{xie2012review, Rayana2015}. 
\cite{barbado2019framework} defined four subsets of user-centric features to analyze Yelp reviews: \textit{personal profile} features (e.g. profiles description), \textit{social interaction} features (e.g., user number of friends), \textit{review activity} features (e.g. number of previous reviews), and \textit{trust information} features (e.g. number of photos posted). Here, they leveraged supervised machine learning techniques to classify fake versus authentic reviews, showing that \textit{review activity} features were the most relevant in terms of classification accuracy. Inherent to our paper, they also described how accounts associated with consistent spamming of fake user-generated content displayed fewer friends, fewer photos posted, and fewer reviews as compared to accounts conducting a genuine activity. 
Strengthening this finding, \cite{luca2016} also confirmed that fake reviews were more likely to be posted by users with less established reputations, as determined by fewer friends and reviews published.

\subsubsection{\textbf{Detection.}}
Human evaluators were found to systematically fail at distinguishing user-generated fake reviews from genuine ones \cite{crawford2015}. For example, \cite{ott2013negative} surveyed members of the general public to detect fake reviews, finding that the best human judge achieved an accuracy of 65\%. Inherently, \cite{ott-etal-2011-finding} found a similar result for the same task amounting to 61\% accuracy. Also, \cite{plotkina2020illusions} and \cite{sun2013synthetic} recorded comparable human performance in similar experimental settings, with humans achieving average accuracy detection rates of 57\% and 52\%, respectively. Such findings demonstrate that humans perform at an accuracy level comparable to random guessing.
In contrast, fake reviews detection, viewed as a binary "\textit{spam} versus \textit{non-spam}" \cite{Jindal2008} or "\textit{fake} versus \textit{non-fake}" \cite{wahyuni2016fake} supervised learning problem, showed promising results. Models such as logistic regression \cite{liu2019, khurshid2019enactment}, naive Bayes \cite{liu2019, zhang2017detecting, heredia2017improving}, random forest \cite{zhang2016online}, and XGBoost \cite{hazim2018detecting} served as valuable benchmarks for more sophisticated models such as deep convolutional neural networks \cite{ren2017neural, zhang2018dri} and recurrent neural networks \cite{zhang2018dri}. Also, large language models (LLM) such as GPT-2 or RoBERTa, depending on the attention mechanism \cite{vaswani2017} have been employed in spam detection tasks \cite{athirai2020gpt2detection, salminen2022creating}. For example, as early as 2011, \cite{ott-etal-2011-finding} achieved detection accuracy rates above 85\% with models such as naive Bayes out of sample. More recently, \cite{salminen2022creating} achieved 97\% F1-score with a RoBERTa transformer in a comparable experimental setting. These findings demonstrate how natural language processing and machine learning techniques outperform human capabilities with a high degree of accuracy and efficiency in terms of identifying patterns and trends in the fake review detection domain.

\section{Methodology} \label{sec:methodology}
%In Figure \ref{fig:pipeline}, it is illustrated the methodology workflow. 

\begin{figure}[h]
  \centering
  \includegraphics[width=\linewidth]{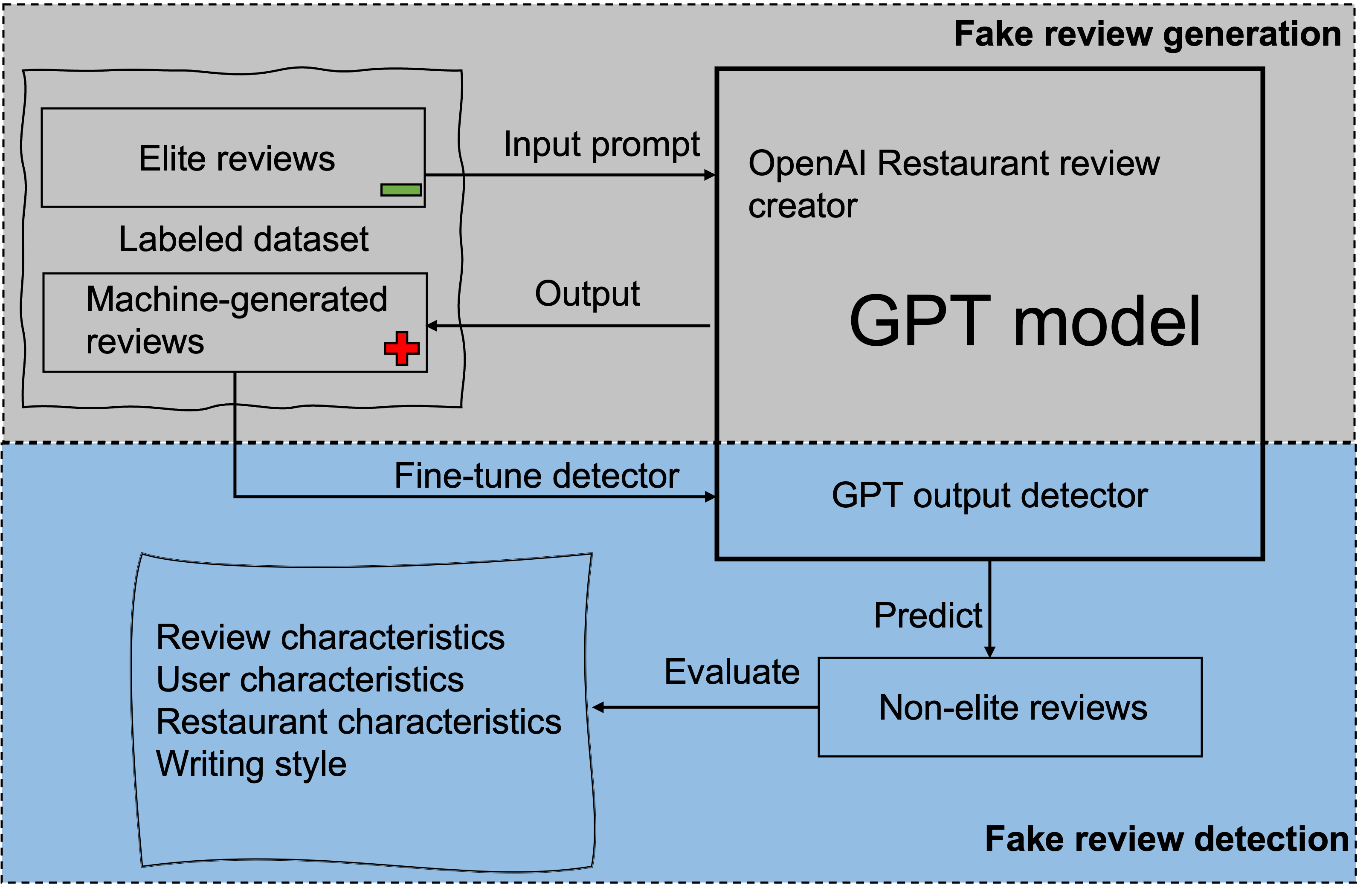}
  \caption{
We leveraged OpenAI's GPT-3 \textit{Davinci} and \textit{Curie} models to create a dataset of fake reviews. Prompts were elite reviews posted by elite users. Therefore, the generated dataset contained equally balanced real elite and fake reviews. To classify them accurately, we fine-tuned a GPT-Neo (GPT-3 equivalent) model, which was  used to predict the probability of non-elite reviews being AI-generated.
  }  \label{fig:pipeline}
  %\Description{Some description}
\end{figure}

In this section, we describe the methodology used. We explain how we: (1) collected data and generated fake GPT-3 reviews, (2) asked human judges and implemented machine learning algorithms to detect them, and (3) inferred and explained the predictions of fake versus non-fake reviews on a set of unverified reviews. 
Figure \ref{fig:pipeline} illustrates the GPT-3 pipeline from fake review generation to detection.
As of 2022, GPT-3 is a state-of-the-art natural language processing model developed by OpenAI that has gained considerable attention due to its impressive performance in a wide range of language-related tasks \cite{brown2020}. Its ability to learn the patterns and structures of language at an unprecedented scale has enabled the model to generate coherent and contextually relevant text that is often indistinguishable from that written by humans. The power and versatility of GPT-3 make it a valuable methodology for both text generation and detection.
It is necessary to highlight that a successful application of GPT-3 models is the recent introduction of ChatGPT, a chatbot that leverages their architecture to engage in human-like conversations and provide support to users in question-answering tasks. However, the wide accessibility and (current) nil usage costs of ChatGPT may also enhance the capabilities of malicious actors to generate fraudulent content to be disseminated on social media platforms. This raises concerns about the potential misuse of GPT-3-based models and underscores the need for vigilant monitoring and control of its applications.

%, with numerous applications ranging from content creation and customer service to language translation and summarization. 
%{\color{red} Its potential for detecting fake or misleading text has important implications for addressing issues of misinformation and disinformation in the digital age. }

\subsection{Data Collection} \label{sec:datacollection} 
We accessed the 2021 to mid-2022 New York City restaurant mobility data from the company SafeGraph (\url{https://www.safegraph.com}) to collect a dataset of restaurants.
%The location of the selection of these restaurants was based on their ability to offer a variety of distinct flavors within an international setting, thereby providing sufficient heterogeneity for our study. 
New York City was selected because it offers a variety of restaurants serving distinct culinary tastes within an international setting, thereby providing sufficient heterogeneity for our study.
Next, we scraped all restaurant-related customer reviews from Yelp.
In total, we collected 447,295 reviews connected to 5,959 restaurants. Each example includes the review text, the date it was posted, the rating, the poster's Yelp elite status, the poster's number of previous reviews, and the poster's number of uploaded photos. Then, we enriched the data by (1) querying the Yelp official API downloading restaurant-related variables (each review was connected to) such as the average rating and the price level, and (2) including the raw number of visits and the normalized number of visits by the total visits (in the New York state) from the original SafeGraph data. 
SafeGraph collects visit data by leveraging various data sources, including GPS signals, Wi-Fi signals, and Bluetooth signals from visitors' mobile devices. The company uses a combination of these signals to determine the temporal location of devices in the physical world, and then aggregates this information to create a comprehensive dataset of visit information. 
%%% 4/4: add some papers that use safegraph as reference 
%It is worth highlighting that SafeGraph is a trustworthy company providing mobility datasets as resources to answer research questions in heterogeneous domains, e.g., public health \cite{chang2021}, impact of mobility restrictions and compliance \cite{charoenwong2020}, disease transmission patterns and epidemiological simulations \cite{he2022, smedemark2022}, and alcohol sales estimation \cite{hu2021}.
%Overall, SafeGraph is a reputable company that has offered reliable mobility datasets to execute research investigations in diverse domains,  
Overall, SafeGraph datasets have been extensively used in diverse research domains, including public health \cite{chang2021}, impact of mobility restrictions and compliance \cite{charoenwong2020}, disease transmission patterns and epidemiological simulations \cite{he2022, smedemark2022}, and estimation of alcohol sales \cite{hu2021}, among others.
%It is noteworthy to emphasize the reliability of SafeGraph as a data source for various research purposes in academic domains, exemplifying its credibility and utility in advancing scholarly investigations.

\subsection{Fake Reviews Generation}
After collecting the main data, we used the OpenAI publicly available GPT-3 API to build a dataset of fake reviews (\url{https://openai.com/api}).
%Given a minimum of very few words as a prompt, a model can generate a fake review of maximum 2,048 tokens. 
As of 2022, four different GPT-3 sub-models could be chosen at different price rates: \textit{Ada} (0.0004\$ / 1K tokens), \textit{Babbage} (0.0005\$ / 1K tokens), \textit{Curie} (0.0020\$ / 1K tokens), and \textit{Davinci} (0.0200\$ / 1K tokens). 
Naturally, the higher the price rate, the more accurate the model instruction-following. A higher price rate also translates into larger model sizes, as measured by the number of parameters, which are not officially disclosed by OpenAI. Simply put, a larger number of parameters leads to increased performance.

%%% 25/03 - TODO: transform the price rate to number of parameters
% ISSUE: OPENAI doeas not officially disclose the number of parameters 
% PEDRO: motivate why we use this models: sota, mass-adoption

%Optionally, an user can tune a temperature value from 0 (less random) to 1 (most random), controlling the randomness in word completion.

We then randomly sampled 12,000 reviews from a total of 92,253 elite reviews (out of 447,295 total scraped reviews) representing 4,994 restaurants, and used the elite-sampled texts as prompts to generate related fake reviews.
Elite reviews are written by elite users, who Yelp thoroughly verifies. According to Yelp, to apply for an elite membership, a user is expected to have consistently posted thoughtful reviews, uploaded beautiful pictures, and up-voted others' reviews. Therefore, we assume that elite reviews are a reliable proxy of information reflecting real customers' opinions. 
As a prompt, we utilized the default template that OpenAI provides for restaurant fake review generation:  
\\

\textit{"Write a restaurant review based on these notes:}

\textit{Name: [Example restaurant name]}

\textit{[Example elite review text]"} 
\\

For each prompt, we randomly selected one model between \textit{Curie} and \textit{Davinci} with equal probability. Also, we randomly sampled the \textit{Temperature} value, a hyper-parameter controlling the randomness of the generated text, from $U\sim(0.3,0.7)$. For general reference, a value of 0 generates a deterministic and repetitive text and 1 vice versa. All the other hyper-parameters were kept as default.
%Finally, we also set the maximum length of generated text as the one of the input prompts. 
Table \ref{tab:dataGeneration} provides three examples with different sentiments.
As a result, the final dataset of 24,000 reviews, equally balanced between elite and fake reviews, was split into 80\% training and 20\% testing for later use. 

\begin{table*}
  \caption{Examples of prompt and GPT-3 generated reviews discriminated by different sentiments.}
  \label{tab:dataGeneration}
  \begin{tabular}{p{1.2cm} p{7.5cm} p{7.5cm}}
    \toprule
    Sentiment & Prompt (Real) & Generated (Fake) \\
    \midrule
    Negative & "Unfortunately I found the soup to be very bland. I'm not sure if it was a one off but I was very disappointed." & "I was really disappointed with Noona Noodles. The soup was very bland and I wasn't sure if it was just a one-off, but I didn't enjoy it at all. I wouldn't recommend it." \\

    Neutral & "Came here for brunch. The shrimp and grits were good so was the okra and molasses corn bread. The owner, Randy, came out to greet the tables. The only issue that almost prevented me from the 4 rating is the long wait for the food. There were several tables complaining. I suggest they give the tables biscuits or rolls to address the hunger while waiting." & "Overall, I had a good experience at Spoonfed NYC. The shrimp and grits were delicious, as was the okra and molasses corn bread. The only downside was the long wait for the food. If they could address the hunger issue, I would have given this restaurant a 4-star rating." \\
    Positive & "So happy there's a dim sum spot near my hood! I only thought Jin Fong only excited in Chinatown. I'm really surprised it has low ratings because every dish I order was really good and service was great! It's a little pricey of what I'm used to of ordering dim sum (\$8 a dish of dim sum). Things I ordered: - Har Gow Shrimp Dumplings- Crab, Shrimp, Spinach Dumplings- Pork and Chicken Soup Dumplings- Shrimp Rice Noodles Rolls- Shrimp and Chive Dumplings" & "If you're in the mood for some delicious dim sum, Jin Fong is the place to go! I was really surprised by how good everything I ordered was - the shrimp rice noodles rolls, shrimp and chive dumplings, and pork and chicken soup dumplings in particular were all delicious. The only downside is that the prices are a little higher than I'm used to paying, but overall it was a great experience." \\
    \bottomrule
  \end{tabular}
\end{table*}

\subsection{Survey Design for Fake Reviews Detection}
% 25/03 - TODO: delete Following... with to assess the human capabiliteis in detecting fake content
%Following \cite{wang2019} methodology with images, 
We ran a human study in which each respondent was asked to select the AI-generated review from a set of review pairs. The surveying strategy consisted of two steps.

Firstly, we sampled and showed 15 review pairs from the training set to "train" the respondents for the task, with each pair containing one human-generated review and one AI-generated review.  
Secondly, we sampled 40 review pairs from the test set and used those as survey questions. 
%%%%%%%% I think this part is getting out of scope %%%%%%%%
Train and test set human-written reviews average about 140 words per review (std 75 words). Here, we picked human-written reviews in the range 140 $\pm$ 40 words away, with reviews exceeding 140 words categorized as \textit{long}, and vice-versa, and paired a comparable length AI-generated review (max 30 words difference). For the questions, we randomly selected 10 same-restaurant long review pairs (\textit{Same-Long}, 140$>$words$>$180), 10 same-restaurant short review pairs (\textit{Same-Short}, 100$<$words$<$140), 10 different-restaurant long review pairs (\textit{Different-Long}), and 10 different-restaurant short (\textit{Different-Short}) review pairs. 
%As for the options, we added a \textit{Cannot decide. I'm unsure} third option not to force participants to randomly guess in case of extreme uncertainty. % rephrase: this is not academic language - MODIFY MODIFY MODIFY
In relation to the response options, a third alternative, \textit{"Cannot decide. I'm unsure"}, was included to enable study participants to indicate their uncertainty instead of having to resort to a random guess. This additional response choice aimed to enhance the accuracy and reliability of the survey data collected by reducing the impact of arbitrary guessing, which may compromise the validity of the study.
All the pairs were randomly spread across the survey form, and 
two questions were converted into attention checks to monitor the respondents' care. 

% send this to results 
The survey was then sent to 90 participants through the Prolific platform, and they were paid about \$8.30 per hour. After removing 10 attempts connected to inattentive answers, we counted 80 valid responses related to 38 questions, totaling 3,040 single-question responses.  
We reported the overall average accuracy and the average accuracy for all four categories. Finally, we conducted a Tukey's HSD test to validate whether such category averages were statistically different.  

%As a result, since the training pairs were intriguing, participants only achieved an average accuracy rate of 57.13\% (13.57\% SD), which is only 7.13\% better than random guessing (50\%). In this calculation, the questions answered as "Cannot decide. I'm unsure" were ruled out, with an average abstention rate of 11.15\%.

\subsection{Automating Fake Reviews Detection with AI} \label{sec:automateDetectionAI}

We deployed machine learning techniques to carry out the same task as above. Practically, we fine-tuned a pre-trained \textbf{GPT-Neo} model \cite{gao2020pile, gpt-neo} to classify fake versus real reviews.

GPT models belong to the family of transformer models, which have become state-of-the-art in natural language processing and computer vision. The reason why transformer models are powerful is that they rely on the attention mechanism \cite{vaswani2017}, allowing the network to focus mainly on the most relevant parts of the input sequence. 
GPT-Neo is designed using EleutherAI's replication of the GPT-3 architecture, which currently is OpenAI's proprietary software. As such, GPTNeo is a scale-up of the GPT \cite{radford2018} and GPT-2 \cite{radford2019} models. 

Practically, we accessed a 125 million parameters pre-trained version from Huggingface (\url{https://huggingface.co/EleutherAI/gpt-neo-125M}), and fine-tuned it with our generated fake restaurant reviews dataset. 
%GPT-2 is a scale-up of the first GPT model \cite{radford2018}, with approximately 10X more parameters and trained on 10X more data. While the last GPT-3 model is OpenAI's proprietary software, GPT-2 is open-source, meaning that official pre-trained versions are accessible to researchers. So, we accessed a 117 million parameters pre-trained version from Huggingface (\url{https://huggingface.co/}), and fine-tuned it with our generated fake restaurant reviews dataset. 
We benchmarked GPT-Neo with other machine learning models such as \textbf{Bi}directional-\textbf{LSTM}, \textbf{L}ogistic \textbf{R}egression, \textbf{N}aive \textbf{B}ayes \cite{manning2008}, \textbf{R}andom \textbf{F}orest \cite{breiman2001}, \textbf{XGB}oost \cite{chen2016}, and \textbf{GPT-2}. We also benchmarked it to the current open-source \textbf{OpenAI}'s RoBERTa model for GPT fake text detection \cite{solaiman2019}.
%Importantly, bidirectional LSTM is a recurrent neural network (RNN) relying on input tokenization, i.e., segmenting the continuous running text into a list of word tokens (numbers), and padding or truncating the sequence to a fixed length to ensure a per-review input equal length. During training, the tokens are sequentially passed to the model. Also, GPT-2 adopts the tokenization rule too. However, the other benchmark models rely on handcrafted features. For this task, we produced a sparse representation of word counts per review. In simple words, a review can be represented as a $n$-long sparse single vector, with $n$ being the number of unique words in the corpora, and with each element $i$ counting the number of each unique word's occurrence. 
We trained with 5-fold cross-validation on the training set for the machine learning models and reported the results on the test set. 
Review texts were represented with a bag-of-words approach, which tokenizes text into individual words and then counts the frequency of those words in each document.
%The hyper-parameters tuning strategies are reported in Appendix A. 
While for the deep learning models (BiLSTM, GPT-2, and GPT-Neo), we extracted another 20\% partition from the training set as validation data since computing 5-fold cross-validation is computationally expensive. 
Here, review texts were tokenized using Byte-Pair Encoding (BPE) \cite{sennrich2015neural}, which is a byte-level data compression algorithm used to segment words into subword units by iteratively merging the most frequently occurring pairs of adjacent bytes.
We trained using the AdamW optimizer default hyper-parameters \cite{loshchilov2017}, using a learning rate of 1e-4, decaying it by a factor of 0.1 every 5 epochs, a batch size of 1, and early-stopping at 10 epochs.
For the GPT-Neo, we computed the optimal classification threshold at each epoch by optimizing Youden's \textbf{J} statistics in the validation set \cite{youden1950}, calculated as the difference between the true positives rate and false positives rate. 
Finally, the best weights and classification threshold were saved, and evaluation was performed on the test set.
For all the models, we reported the accuracy score, precision score, recall score, and F1 score.

\subsection{Inference on Non-Elite Reviews} \label{sec:inference}
With the best GPT-Neo classifier, we performed inference on the unverified non-elite reviews, determining the probability of each one being fake. Importantly, reviews posted before 2020 (included) were dropped, since GPT-3 models were beta-released in 2020. Here, we hypothesized that GPT-3-based AI crowdturfing campaigns were implemented in the last two years, given the GPT-3 API's easy accessibility and low usage costs. This operation reduced the inference dataset to 131,266 non-elite reviews. Importantly, Yelp implements a proprietary algorithm to flag and filter fake reviews \cite{luca2016reviews}. In this paper, we performed inference on reviews that had already passed the Yelp filtering system.

Each example review incorporates a \textit{review}-based variable, i.e. the review rating (\textit{Rating}), distributed as a 1 to 5 Likert scale; \textit{user}-based variables, i.e. the user's number of friends (\textit{\#Friends}), the user's aggregated number of previously posted reviews (\textit{\#Reviews}), and the user's overall number of posted photos (\textit{\#Photos}); and \textit{restaurant}-based variables, i.e. the restaurant's average rating from all the reviews (\textit{AvgRating}), the price level (\textit{PriceLevel}), i.e. the average price per person as “\$”:
under \$10, “\$\$”: \$10–\$30, “\$\$\$”: \$31–\$60 and “\$\$\$\$”: over \$60, 
the overall number of restaurant reviews posted by customers (\textit{\#RestReviews}), the chain status (\textit{ChainStatus}), computed adopting Zhang \& Luo (2022) approach \cite{zhang202}, which counts the number of unique restaurant names in the dataset, and assigns those appearing more than five times as belonging to a restaurant chain (e.g. McDonald's, Starbucks, Burger King, etc.), the number of customer visits between 2021 and mid-2022 (\textit{\#Visits}), and the normalized number of visits (\textit{NormVisits}), multiplied by 1,000 for easier readability. A summary statistics is provided in Table \ref{tab:variablesSummaryStats}.

Afterward, classification was performed with a sensitivity analysis approach at the [.5, .6, .7, .8, .9, .99 and \textbf{J}] classification thresholds. For each threshold, we separated predicted fake versus non-fake reviews, and performed ANOVA for each aforementioned variable to inspect differences across the two predicted categories.
This methodology was adopted because the labels about whether non-elite reviews were AI-generated were not available. Thus, different thresholds were tested to examine the sensitivity and robustness of predictions.

%{\color{red} This approach is motivated by the fact that we do not have the ground truth labels of each review, so we test with different thresholds as proxies for different degrees of confidence.}

% REPHRASE: motivate the sensistivity analysis saying that we dont have the ground truth, so we test with different threshold with different degrees of confidence 

\begin{table}
  \caption{Summary statistics of non-elite reviews related variables.} \label{tab:variablesSummaryStats} 
  \begin{tabular}{cccl}
    \toprule
    Name & Category & Mean & Std\\
    \midrule
    Rating & Review & 3.98 & 1.44 \\
   \#Friends & User & 70.71 & 174.22\\
     \#Reviews & User & 43.25 & 129.56  \\
     \#Photos & User & 63.90 & 722.89 \\
     AvgRating & Restaurant & 3.96 & 0.49 \\
     PriceLevel & Restaurant & 2.23 & 0.67 \\
     \#RestReviews & Restaurant &  747.47 & 1045.94 \\
     \#Visits & Restaurant & 2993.03 & 5835.29\\
     NormVisits & Restaurant & 0.19 & 0.37 \\
     ChainStatus & Restaurant & 0.14 & 0.35 \\ 
  \bottomrule
\end{tabular}
\end{table}

%Unless differently stated, the ANOVA significance threshold $\alpha$ was set to 0.05.
% optional 
%Namely, if more than 5 restaurants with the same name, but at different locations, were present in the data, then the restaurant is a chain (e.g. McDonald's, Burger King, KFC, etc.). Recognizing chain restaurant is easy, because Yelp uses an alias naming convention “name–location–number” to uniquely identify specific restaurants. Then, for each unique "name", we collected the related "number"(s), and if the maximum was greater than 5, then the restaurant sharing the same name were labeled as chain. 

\subsection{Writing Style: Explaining the Predictions} % writing style
In our context, writing style refers to how a textual review is constructed by the writer, sentence-by-sentence, and word-by-word. Although it is hard to distinguish at first impact, we believe that humans and AI have different writing styles, with the latter being more repetitive, more predictable, and less sophisticated than the former. 

We considered three classes of metrics to evaluate the writing style of each non-elite review: \textit{perplexity}-based, \textit{readability}-based, and \textit{sentiment}-based metrics. \textit{Perplexity}-based metrics include \textit{Perplexity} (\textit{PPL}) and \textit{Textual Coherence} (\textit{TC}). 

\begin{equation} \label{eq:ppl}
    PPL(X) = \exp \left[ -\frac{1}{t} \sum_1^{t} \log p(x_i | x_{<i}) \right]
\end{equation}

As in Equation \ref{eq:ppl}, \textit{PPL} is defined as the exponential average negative log-likelihood of a sequence of words $w_i, w_{i+1} \dots w_{i+t} $. In simple terms, it measures the conditional probability that each word follows its preceding one. As of 2022, \textit{PPL} is one of the most widely adopted metrics to evaluate the accuracy of language models. Generally, a low \textit{PPL} score implies better grammar text correctness and cohesion.

Then, by breaking a review into a sequence of sentences, we introduce the concept of \textit{Textual Coherence} (\textit{TC}). \textit{TC} is defined as the presence of semantic relations among sentences. In simple words, given a corpus containing a set of sentences that when viewed independently convey a valid meaning, if by reading them sequentially no meaning is conveyed, then the corpus is not coherent.
To measure \textit{TC}, we deployed the \textit{Zero-Shot Shuffle Test} \cite{laban2021, wang2014}. 
Namely, we split each review into single sentences and generated all the possible sentence permutations. We scored each permutation with the \textit{PPL} as in Equation \ref{eq:ppl} and subtracted the original perplexity score, obtaining a per-review set of perplexity changes, which we averaged to compute \textit{TC}. 
It is important to highlight that some reviews contained a large number of sentences, leading to high computational costs in generating permutations. Mathematically, given a set of $n$ sentences (population), and a subset of $r$ sentences to be chosen from $n$, it is possible to generate all the permutations without repetition $P(n,r)=\frac{n!}{(n-r)!}$. In our problem, $n=r$, meaning that $P(n,r)=n!$, which incurs an expensive computational cost $O(n!)$. To mitigate that, we sampled $s$ sentences from each review such that $s=min(n, 5)$. 
We chose 5 as a maximum threshold because such value represents the median number of sentences per review, and because $5!$ per-review permutations are still computationally feasible to be processed. To calculate both \textit{PPL} and \textit{TC}, we adopted a general purpose pre-trained 125 million parameters GPT-Neo. 

As for \textit{readability}-based metrics instead, we considered the following metrics: \textit{Automated Readability Index} (\textit{ARI}) \cite{senter1967},  \textit{Number of Difficult Words} (\textit{\#DW}), and \textit{Readability Time} (\textit{RTime}).
\textit{ARI} is one of the most widely adopted readability indices to evaluate the readability of a given text. Also, it has already been adopted in other studies evaluating the readability of online reviews (e.g. \cite{hu2012, harris2012detecting}). 

%In the same domain, also the \textit{Gunning-FOG} index \cite{gunning1969} has been used (e.g. \cite{fang2016}). However, given that it requires a minimum of 100 words, its application to our case would not be stable as not all the reviews exceed that minimum number of words.

% 25/03 - TODO: part on gunning fog is too loose. rephrase it - MAYBE DELETE IT 

\begin{equation} \label{eq:ari}
    ARI= 4.71 \frac{\#Chars}{\#Words} + 0.5 \frac{\#Words}{\#Sentences} - 21.43
\end{equation}

As in Equation \ref{eq:ari}, \textit{ARI} decomposes the text into basic structural elements such as the number of characters (\textit{\#Chars}), number of words (\textit{\#Words}), and number of sentences (\textit{\#Sentences}). Unlike other readability indices, the main advantage of \textit{ARI} is that it relies on the number of characters per word and not on the number of syllables per word, therefore being more accurate to calculate for a computer. Also, the interpretation of \textit{ARI} is straightforward, as its output produces an approximated representation of the US-grade education level needed to understand the text. For example, an \textit{ARI} of 9.2 indicates that a 9th-grade student can understand the text. Simply put, the higher the \textit{ARI} score, the higher the difficulty in text comprehension for an average interlocutor. 
Next, \textit{\#DW} is the count of difficult words present in a text. By looking at the \textit{Dale-Chall Word List} \cite{dale1948}, which contains approximately 3,000 familiar words known by an average 5th-grade student, if a word is not present in the list, then it is labeled as difficult. 
Then, \textit{RTime} was computed by following \cite{demberg2008}, who found that each text character needs an estimated average of 14.69 milliseconds to be digested by the reader. %Thus, the reading time in seconds is simply computed by multiplying the number of characters by 14.69\textit{ms}.
Finally, the only \textit{sentiment}-based metric is the SiEBERT \textit{Sentiment} score \cite{hartmann2022}. SiEBERT is based on a RoBERTa architecture and fine-tuned on 15 different datasets. Its output ranges from -1 (negative) to +1 (positive). 

To sum up, we scored each review with the above-mentioned \textit{perplexity}-based, \textit{readability}-based, and \textit{sentiment}-based metrics. Afterward, for each metric, we performed ANOVA to inspect differences across the predicted fake versus real reviews with the same methodology mentioned in Section \ref{sec:inference}.

\section{Results} \label{sec:results}

%\cite{Cohen:2007:DEC:1219092.1219093}

\subsection{Human Evaluations versus Model Evaluations} \label{sec:humanEvalVsModelEval}
Surveyed people from the general public only attained an accuracy score of 57.13\% (std 13.57\%), meaning that humans are only 7.13\% better than random guessing (=50\%). In addition, we recorded an abstention rate (i.e. selecting the \textit{"Cannot Decide. I'm unsure option"}) of 11.15\% (std 12.66\%). 
In Table \ref{tab:survey}, we report the Tukey-HSD results across the different categories. 
\begin{table}
  \caption{Tukey-HSD test results across categories. In brackets the category unweighted by abstention rate accuracy averages in percentage. "Different" stands for different restaurants, and "Same" vice versa. "Long" stands for a long review, and "Short" vice versa.  *\textit{p}<.05, **\textit{p}<.01, ***\textit{p}<.001.} \label{tab:survey}
  \begin{tabular}{cccl}
    \toprule
    Category 1 & Category 2 & MeanDiff \\
    \midrule
    Different-Long (50.76) & Different-Short (55.11) & 4.35 \\
    Different-Long (50.76) & Same-Long (61.39) & 10.63* \\
    Different-Long (50.76) & Same-Short (61.93) & 11.17** \\
    Different-Short (55.11) & Same-Long (61.39) & 6.28 \\
    Different-Short (55.11) & Same-Short (61.93) & 6.82 \\
    Same-Long (61.39) & Same-Short (61.93) & 0.54 \\
     \bottomrule
  
\end{tabular}
\end{table}
Here, we did not discover significant differences across categories, except for (1) \textit{Different-Long} versus \textit{Same-Long} (10.63, \textit{p}<.05), and (2) \textit{Different-Long} versus \textit{Same-Short} (11.17, \textit{p}<.01) pairs. Apparently, for long reviews, humans are more accurate in distinguishing fake content when the review is associated with the same restaurant. 
However, with these results, we concluded that humans are not generally capable of distinguishing real versus fake content. 
Conversely, machine learning algorithms attained significantly better performance than human evaluators. In Table \ref{tab:modelResults} we provide the classification report of the classifiers. 

\begin{table}
  \caption{Models classification report on test set ranked by F1-score.} \label{tab:modelResults}
  \begin{tabular}{ccccl}
    \toprule
    Model  & Accuracy & Precision & Recall & F1-score \\
    \midrule
        OpenAI & 76.78 & 84.87 & 64.98 & 73.60 \\
        NB & 83.48 & 93.62 & 71.72 & 81.22  \\
        RF & 82.39 & 84.61 & 79.01 & 81.72 \\
        XGB & 83.18 & 86.59 & 78.38 & 82.28  \\
        LR & 85.07 & 87.00 & 82.34 & 84.61  \\
        BiLSTM  & 93.71 & 93.09 & 93.09 & 93.09  \\
        GPT-2 & 94.63 & 94.57 & 94.65 & 94.61  \\
        %GPT-2@J & 94.88 & \textbf{95.98} & 93.64  & 94.80  \\  
        GPTNeo & 95.21 & 94.74 & \textbf{95.70} & 95.22 \\
        \textbf{GPTNeo@J} & \textbf{95.51} & \textbf{95.80} & 95.15 & \textbf{95.48}  \\
  \bottomrule
\end{tabular}
\end{table}

We observed that the current OpenAI's fake text detector benchmark improved human evaluators' accuracy by 19.65\%, meaning that machines are more suitable for performing the task. 
This claim was strengthened by looking at the performance of standard machine learning algorithms. In particular, LR ranked as the top performer among those with an accuracy score and F1-score of 85.07\% and 84.61\%, respectively.  As for deep learning models, GPT-Neo models ranked as top performers. Specifically, the GPT-Neo maximizing accuracy after calculating the Youden's-J statistics as the optimal classification threshold in the validation set (GPT-Neo@J) achieves the best performance. Convergence occurred at the 2nd epoch, with optimal $J^{*}$=.5708. 
Overall, GPT-Neo@J significantly outperforms human evaluators and OpenAI's benchmark by 38.38\% and 18.73\%, respectively. With this finding, we applied the optimized GPT-Neo@J model for inference on the unverified non-elite reviews.
%{\color{red} Finally, we also calculated the Fleiss' Kappa index to measure the inter-rater reliability agreement among humans (.047) and among machines (.73). A score of 1 indicates perfect agreement and 0 vice versa. These results showed that humans could not substantially agree with each other, while machines considerably could.} 
% 25/03 - TODO: rephrase the part of Fleiss Kappa. it is disconnected 
% 31/03 -> MAYBE PUT IN APPENDIX 

\subsection{ANOVA Results}
Unless differently specified, ANOVA results are discussed at the .05 significance level and at the optimized classification threshold $J^{*}$ found in Section \ref{sec:humanEvalVsModelEval}. 
In Table \ref{tab:ANOVAResults}, we provide a per-variable summary. 

%Overall, 8.48\% of reviews were predicted as fake out of a total of 131,266 non-elite reviews posted from 2021 onward. In Figure \ref{fig:percentagePredictedFake}, we show how this percentage depreciates by lowering the threshold $t$. \color{red} For example, at $t$=.99, only .10\% of reviews were filtered out as AI-generated.
% 25/03 - TODO -> REPHRASE this 

\begin{figure}[h]
  \centering
  \includegraphics[width=\linewidth]{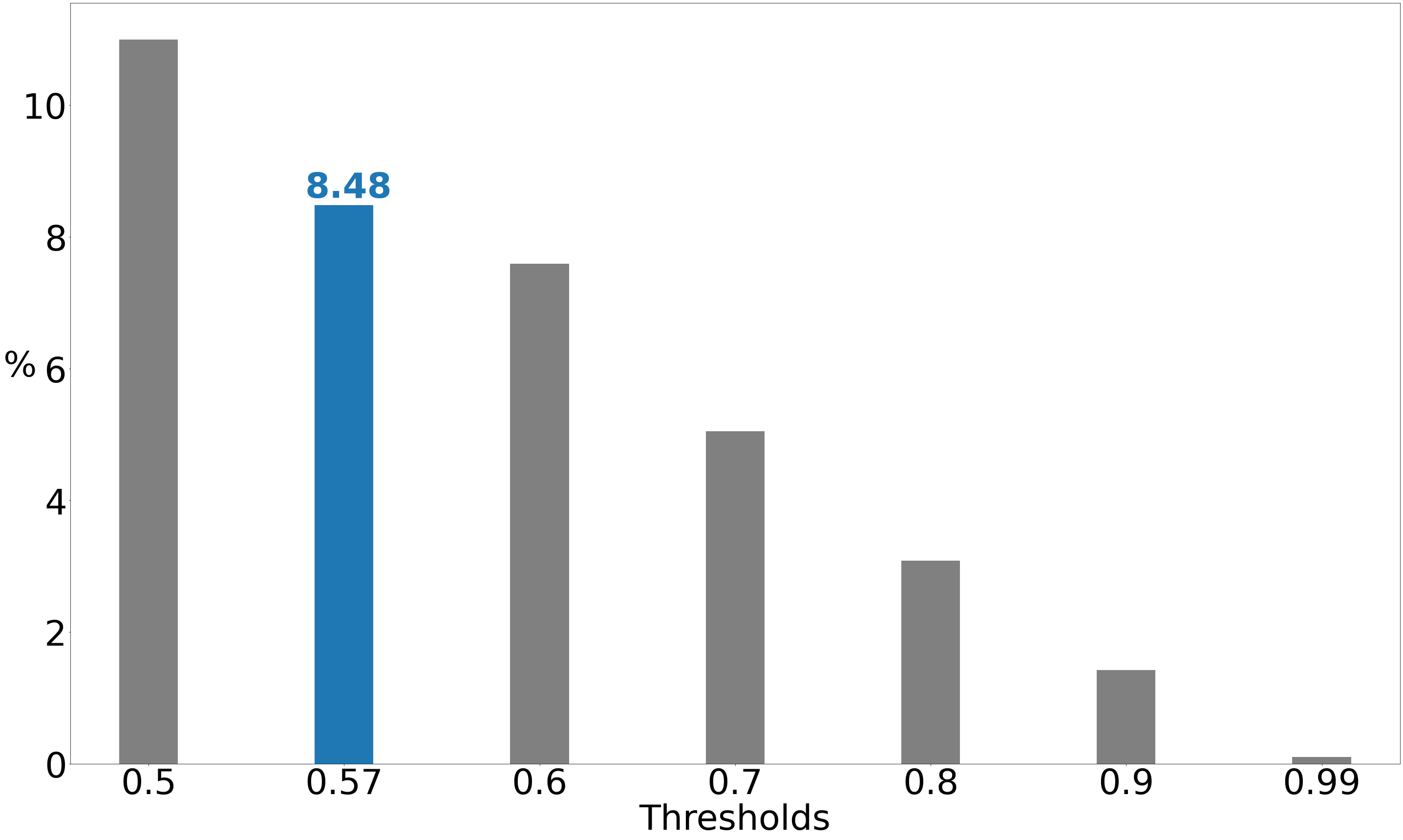}
  \caption{Percentage of predicted fake reviews at each threshold. In blue, highlighted the $J^{*}$=.5708 optimal threshold.}  \label{fig:percentagePredictedFake}
  \Description{Some description}
\end{figure}

% -> REPHRASED
Out of a total of 131,266 non-elite reviews posted from 2021 onwards, 8.48\% were predicted as fraudulent. As demonstrated in Figure \ref{fig:percentagePredictedFake}, this percentage decreases as the threshold $t$ is lowered. For instance, at a threshold of $t$=.99, only .10\% of reviews were detected as AI-generated.

%Also, we break this section by presenting Review-based and User-based, Restaurant-based, and lastly Writings-style in separated subsections.

%%%%%%% P VALUES STARS rule %%%%%%%
%If a p-value is less than 0.05, it is flagged with one star (*). If a p-value is less than 0.01, it is flagged with 2 stars (**). If a p-value is less than 0.001, it is flagged with three stars (***). 

\begin{table} 
  \caption{
  Summary of ANOVA results at  $J^{*}$=.5708 classification threshold for the average predicted human (real) and AI (fake). *\textit{p}<.05, **\textit{p}<.01, ***\textit{p}<.001.}\label{tab:ANOVAResults}
  \begin{tabular}{ccccl}
    \toprule
    Name & Category & Human & AI & F-statistic\\
    \midrule
     Rating & Review & 3.94 & 4.37 & 914.91*** \\
    \#Friends & User & 71.19 & 65.60 & 10.45**\\
     \#Reviews & User & 44.21 & 32.84 & 78.46*** \\
     \#Photos & User & 66.81 & 32.49 & 22.96***  \\
    AvgRating & Restaurant & 3.97 & 4.00 & 41.84*** \\
     PriceLevel & Restaurant & 2.24 & 2.22 & 3.52 \\
     \#RestReviews & Restaurant & 743.68 & 788.39 & 18.61*** \\
     \#Visits & Restaurant & 3004 & 2866 & 5.22* \\
     NormVisits & Restaurant & 0.19  & 0.18 & 5.18*  \\
     ChainStatus & Restaurant & 0.14 & 0.14 & 0.04  \\ 
    Perplexity  & Writing & 78.70 & 83.38 & 14.01*** \\ %% CAREFUL
     Coherence  & Writing & 25.31 & 21.20 & 3.33\\  %% CAREFUL
     ARI & Writing & 7.05 & 6.82 & 13.35***\\
     \#DW & Writing & 10.70 & 6.76 & 1847.82***\\
     RTime & Writing & 5.21 & 2.87 & 2532.22*** \\
     Sentiment & Writing & 0.47 & 0.71 & 779.17*** \\
     
  \bottomrule
\end{tabular}
\end{table}

% EDITED TO GPTNeo
\subsubsection{\textbf{Review-based and User-based.}}
\begin{figure}[h]
  \centering
  \includegraphics[width=\linewidth]{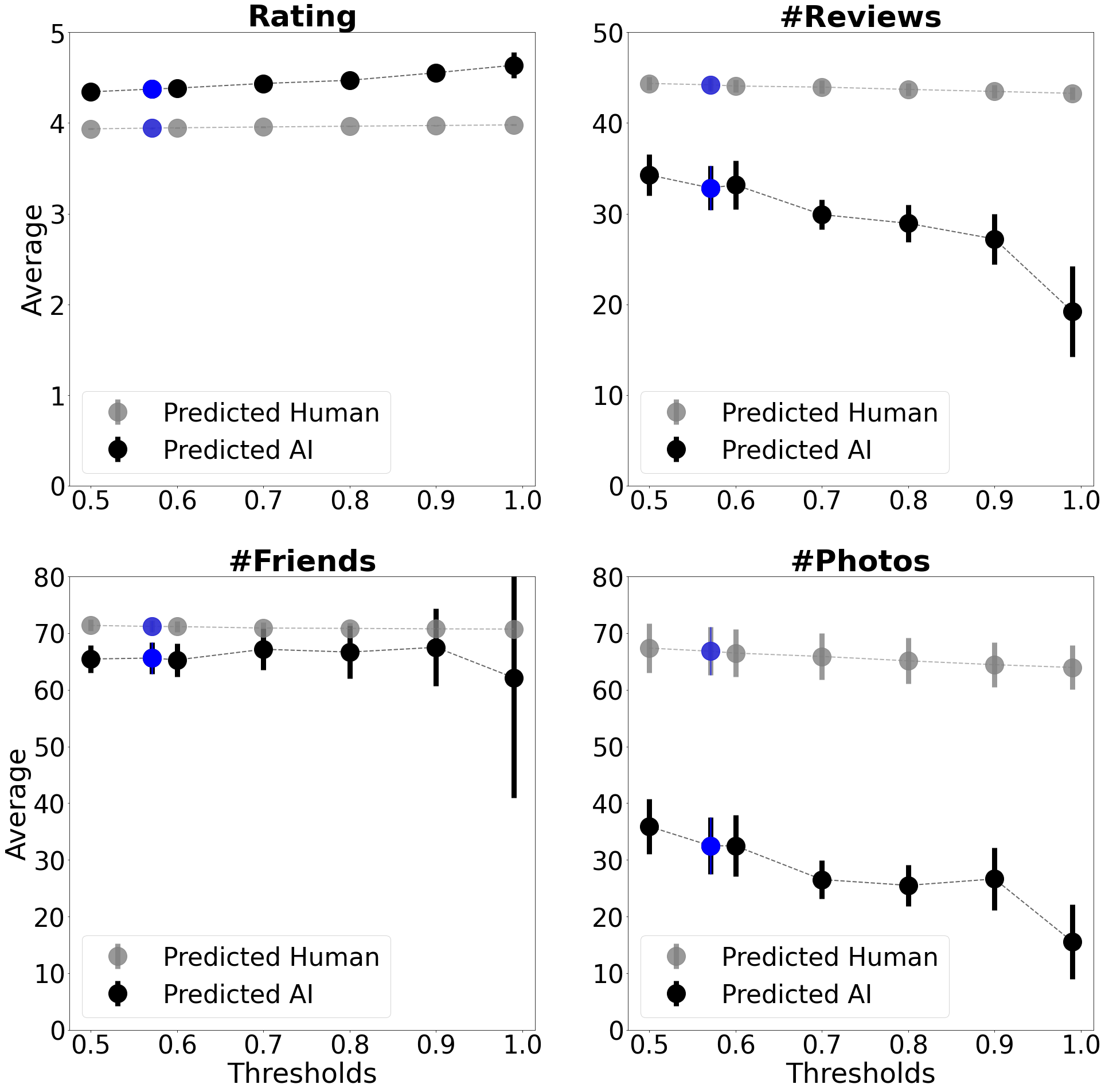}
  \caption{Sensitivity analysis of review-based and user-based variables. In blue, highlighted the $J^{*}$=.5708 optimal threshold value. Error bars are the confidence intervals at the .05 significance level.} \label{fig:reviewAndUserSensitivity}
  \Description{Some description}
\end{figure}

For the review \textit{Rating}, and the users' \textit{\#Reviews}, \textit{\#Friends} and \textit{\#Photos} all differences across predicted fake AI-generated and predicted real reviews were statistically significant. Here, reviews classified as fake were given a higher average star \textit{Rating} (+.43, $p$<.001). 
As for user-based variables, reviews classified as fake were posted by users with a lower average number of \textit{\#Friends} (-5.59, $p$<.01), a lower average number of previously posted \textit{\#Reviews} (-11.37, $p$<.001), and a lower average number of previously posted \textit{\#Photos} (-34.32, $p$<.001). In Figure \ref{fig:reviewAndUserSensitivity}, we show a sensitivity analysis considering other thresholds $t$ of classification.
%Notably, at $t=.99$, \textit{Rating} and \textit{\#Reviews} display statistically significant higher divergences of .66 ($p$<.001) and -24.06 ($p$<.05), respectively. %Whereas, \textit{\#Friends} and \textit{\#Photos} lose statistical significance.
%In the evaluation of the differences between predicted fake AI-generated reviews and predicted genuine reviews, the variables \textit{Rating}, \textit{\#Reviews}, \textit{\#Friends}, and \textit{\#Photos} all exhibit statistically significant disparities. Specifically, reviews identified as fraudulent received a higher mean star rating (increase of .43, $p$ < .001).
%Regarding user-based variables, counterfeit reviews were associated with users who had a smaller average number of friends (decrease of 5.59, $p$ < .01), a lower mean count of previously submitted reviews (decrease of 11.37, $p$ < .001), and a reduced average number of previously uploaded photos (decrease of 34.32, $p$ < .001). Figure \ref{fig:reviewAndUserSensitivity} illustrates a sensitivity analysis that considers alternative classification thresholds, denoted as $t$.
Remarkably, when the threshold $t$ is set to .99, both \textit{Rating} and \textit{\#Reviews} exhibited greater statistically significant differences, with disparities of +.66 ($p$ < .001) and -24.06 ($p$ < .05), respectively. In contrast, the variables \textit{\#Friends} and \textit{\#Photos} no longer demonstrated statistical significance at this threshold.

\subsubsection{\textbf{Restaurant-based.}}
\begin{figure}[h]
  \centering
  \includegraphics[width=\linewidth]{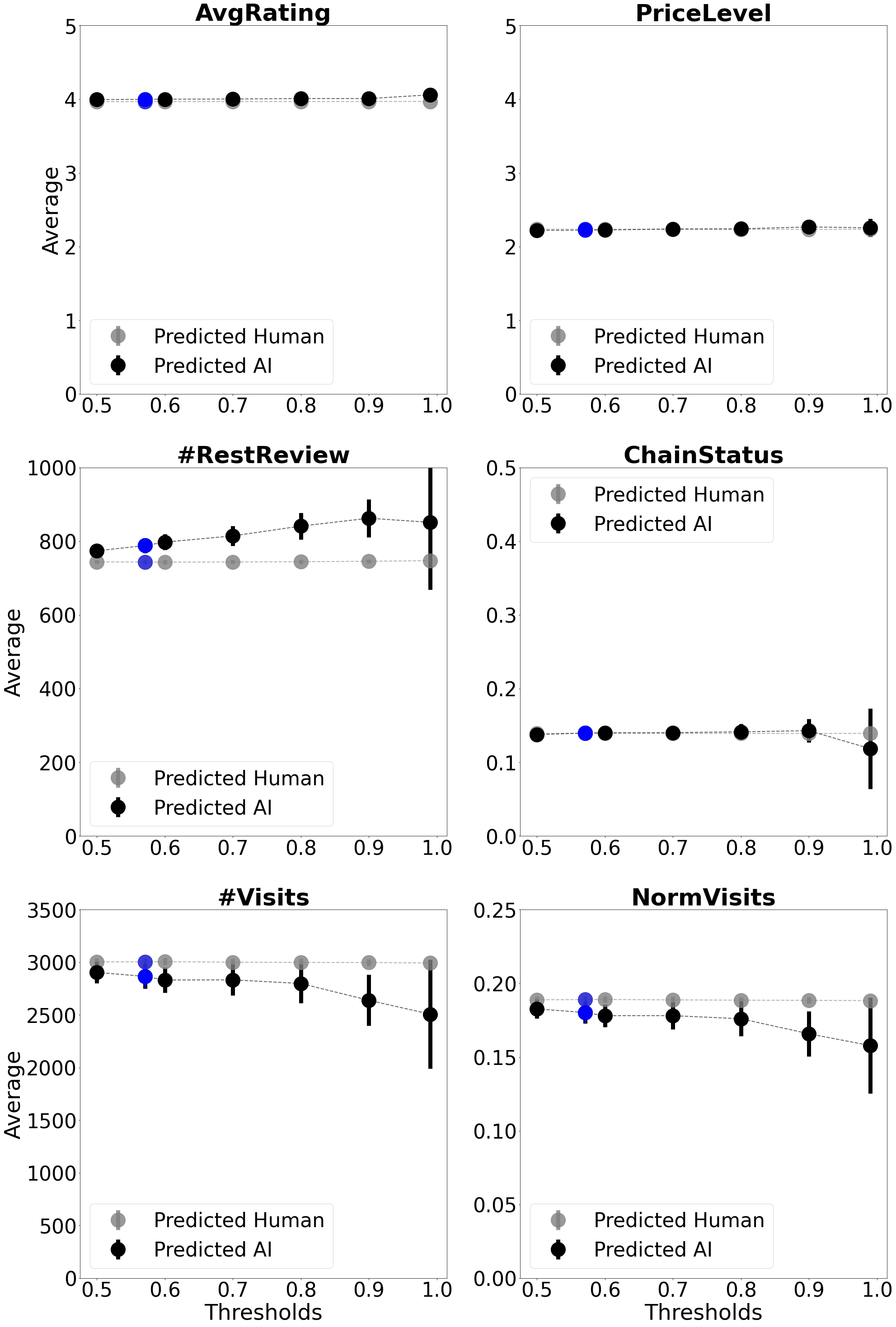}
  \caption{Sensitivity analysis of restaurant-based variables. In blue, highlighted the $J^{*}$=.5708 optimal threshold value. Error bars are the confidence intervals at the .05 significance level.} \label{fig:restaurantSensitivity}
\end{figure}

We observed that predicted fake reviews were associated with restaurants with a higher \textit{AvgRating} (+.03, $p$<.001). However, given the minimal difference, we acknowledge the modest practical implications that such a result may bring about. 
Then, we documented statistical significance concerning \textit{\#RestReviews}. Here, predicted fake reviews were connected to restaurants that displayed a greater average number of reviews available (+44.71, $p$<.001). The opposite behavior was observed for the average \textit{\#Visits}, in which predicted fake reviews were linked to restaurants that received fewer customer visits from 2021 to mid-2022 (-138, $p$<.05). This result was strengthened by \textit{NormVisits} (-.01, $p$<.05). 
Finally, no significant differences were noticed for the \textit{ChainStatus} and the \textit{PriceLevel} ($p$>.05). In Figure \ref{fig:restaurantSensitivity}, we show a sensitivity analysis considering other thresholds $t$ of classification.
%Here, we observed an increasing trend for \textit{\#RestReviews}. At $t$ = 0.99, predicted fake reviews were associated with restaurants displaying a higher number of reviews (+121.38, $p$<.001). The other variables did not exhibit significant trends.

%%%%%% TO EDIT 
\subsubsection{\textbf{Writing Style.}}
\begin{figure}[h]
  \centering
  \includegraphics[width=\linewidth]{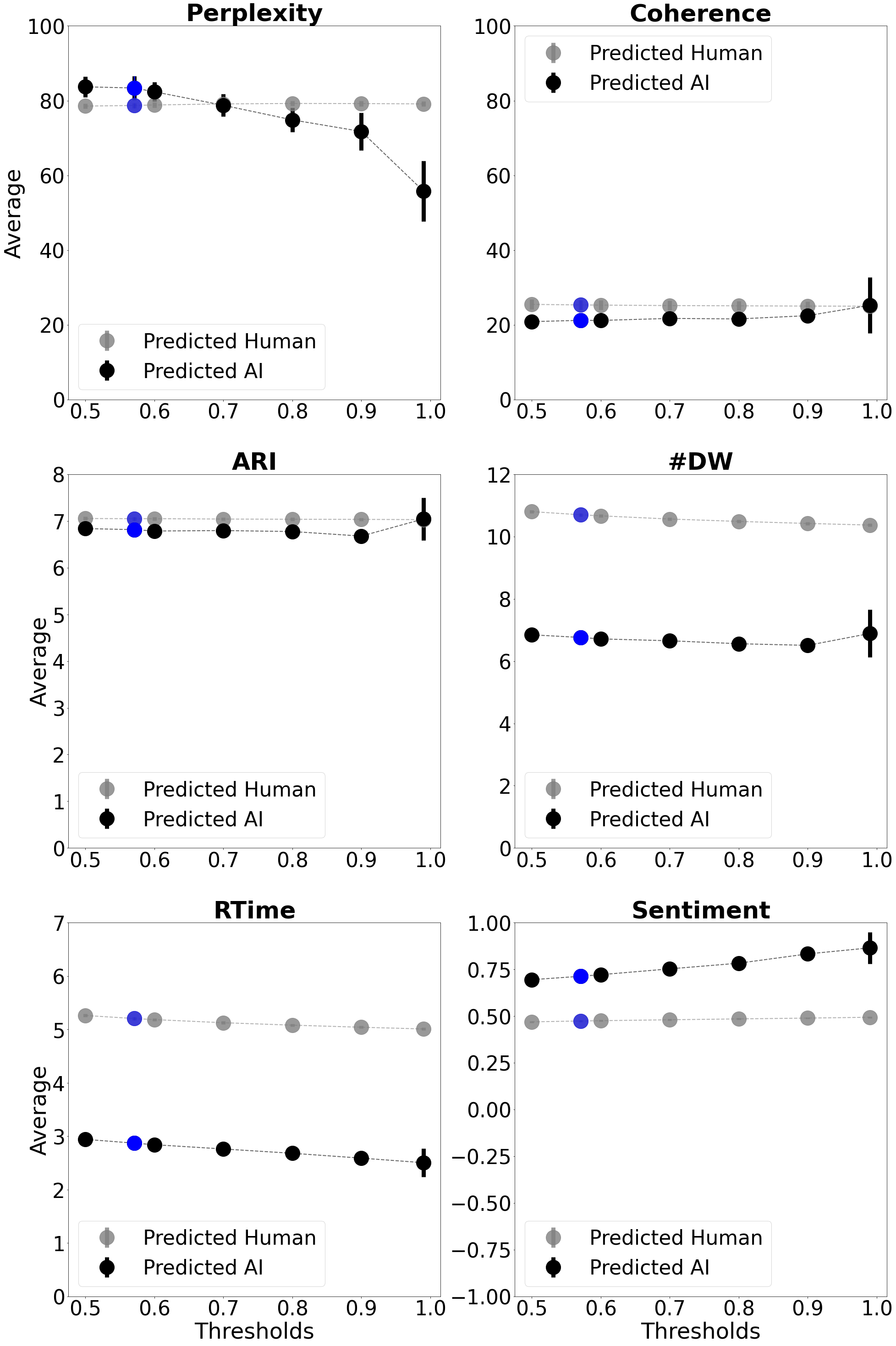} % plot6 is the 3x2
  \caption{Sensitivity analysis of writing-style variables. In blue, highlighted the $J^{*}$=.5708 optimal threshold value. Error bars are the confidence intervals at the .05 significance level.} \label{fig:writingSensitivity}
  \Description{Some}
\end{figure}

AI-generated fake reviews showed a higher average \textit{Perplexity} (+4.68, $p$<.001). However, \textit{Perplexity} of predicted AI-generated fake reviews also displayed a statistically significant downtrend when increasing the classification threshold $t$ ($p$<.05), eventually scoring lower than for predicted human-generated reviews from $t$>.7.
%, implying that they were less grammatically correct and less predictable compared to the human-generated ones. 
Instead, when considering \textit{Textual Coherence}, i.e. defined as the change in perplexity between sentence-shuffled documents, no statistical significance was observed ($p$>.05).
Figure \ref{fig:writingSensitivity} shows the sensitivity analysis.

%This result was further strengthened by Figure \ref{fig:writingSensitivity}, which shows a sensitivity analysis at sequential thresholds. 
%For AI-generated content, a higher $t$ led to a statistically significant lower average perplexity but lower average coherence, i.e. a positive perplexity change. Moreover, divergences from human-generated appreciated, quantified at -21.97 ($p$<.001) and +1.51 ($p$<.001) at $t$=0.99, respectively.

As for \textit{readability}-based metrics, predicted AI-generated fake reviews were discovered to be more readable and less difficult to comprehend compared to the human-generated ones. Both average \textit{ARI} and average \textit{\#DW} scored lower for AI-generated content,  (-.23, $p$<.001) and (-3.94, $p$<.001), meaning that such content can be understood by a wider audience. Also, predicted AI-generated fake reviews were faster to read by 2.34 seconds ($p$<.001).
%To conclude, we did not observe neither increasing nor decreasing trends by changing the threshold of classification here. However, at each $t$, all differences were statistically significant.
Lastly, for \textit{Sentiment}, we observed that predicted AI-generated reviews have a more positive tone (+.24, $p$<.001). Here, predicted human text averaged to a medium positive polarity close to .5 at each $t$; while, predicted AI-generated text displayed a statistically significant positive trend, ending up at .87 at $t$=.99 ($p$<.001).
%Interestingly, at $t$=.99, the sentiment polarity averaged to .87 ($p<$.001).

\section{Discussion} \label{sec:discussion}

\begin{table}
  \caption{Summary of the findings from extant literature related to \textbf{UGC fake reviews} extend by this research to the realm of AIGC fake reviews.} \label{tab:extended_studies}
  \begin{tabular}{p{1.5cm} p{0.5cm} p{2.5cm} p{2.5cm}}
    \toprule
    Variable & Study & UGC Prior Results & AIGC Results (ours) \\
    \midrule
    Rating & \cite{luca2016} & Bimodal distribution. Spikes at 1 star and 5 stars. & Fake more positive.  \\

    Rating & \cite{lappas2016impact}  & Fake more positive. & Fake more positive. \\

    \#Friends & \cite{barbado2019framework} & Fake fewer friends. &  Fake fewer friends.\\
    \#Friends & \cite{luca2016} & Fake fewer friends. &  Fake fewer friends. \\
    \#Photos & \cite{barbado2019framework} & Fake fewer photos. & Fake fewer photos.\\
    \#Reviews & \cite{barbado2019framework} & Fake fewer reviews. &  Fake fewer reviews. \\
    \#Reviews & \cite{luca2016} & Fake fewer reviews. &  Fake fewer reviews. \\
    %AvgRating & & & \\
    ChainStatus & \cite{luca2016} & Chain display less fake content. & No statistical difference. \\
    \#RestReviews & \cite{luca2016} & Stronger incentives to post fake reviews when few reviews displayed.  & Fake more reviews.\\
    %Perplexity & \cite{mitrovic2023chatgpt} & Less & \\
    ARI & \cite{harris2012detecting} & Fake less complex. & Fake less complex.\\
    ARI & \cite{Yoo2009} & Fake more complex. &  Fake less complex. \\
    Sentiment & \cite{Yoo2009} & Fake more positive. & Fake more positive.\\
    Sentiment & \cite{banerjee2014theoretical} & Fake more positive. & Fake more positive.\\
    Sentiment & \cite{liu2018unified} & Fake more polarized. & Fake more positive. \\
    
  \bottomrule
\end{tabular}
\end{table}

% 25/03 - IN DISCUSSION: 
% discuss why AI generated text is more easier to read. 
% FIND JUSTIFICATION 4 EVERY FINDING 

Aligned with findings from prior studies in fake reviews identification \cite{salminen2022creating, ott-etal-2011-finding, sun2013synthetic}, we described how human evaluators systematically fail at detecting GPT-3 AI-generated content (AIGC) in the domain of restaurant reviews. On the contrary, machine learning models significantly achieve superior performance (+38.38\% accuracy). We have also shown that optimizing Youden's J statistic in the validation set can further improve prediction accuracy (+.3\%). Such disparity in performance between humans and machines could be possibly attributed to human cognitive limitations at detecting patterns from large-scale unstructured data \cite{kahneman1972subjective}.
%{\color{red} .} % talk about human cognitive limitations
%{\color{red} In addition, in line with \cite{salminen2022creating}, as measured by the Fleiss' Kappa index, we have documented that machines agree with each other more than humans do in detecting fake content, implying that machines can establish a general uniform consensus to successfully perform in the task as opposed to humans, who fail to detect patterns from large-scale unstructured data because of cognitive limitations \cite{kahneman1972subjective}.} %%% MAYBE PUT IN APPENDIX 
Prior studies that leveraged machine learning techniques reported filtering out user-generated fake reviews at estimated rates around 15\% \cite{luca2016, sussin2012consequences, lappas2016impact}. In our sample of customer reviews that already passed the filtering system of social media (Yelp), we further documented that up to 8.48\% of them were fake reviews generated by AI.
%This finding leads us to hypothesize that fake human-generated content is currently more prevalent than fake AI-generated content. % is there a more recent study on filtering UCG and stating its percentage?
We then explored how fake AI-generated reviews differentiate from human-generated reviews across several associated dimensions. 
In Table \ref{tab:extended_studies}, we summarize extant literature findings on user-generated fake reviews compared to our findings on AI-generated fake reviews.
Firstly, we observed that predicted AI-fake reviews score a higher average \textit{Rating} compared to the human-generated ones (+.43, $p$<.001). This finding is congruent with conclusions from \cite{luca2016}, who documented that fake reviews have a bimodal distribution with spikes at 1 and 5 stars, and from \cite{lappas2016impact}, who singled out 56\% of positive reviews out of 15,000 hotel fake reviews on Yelp.
% 31/03 -> justify 
Extant literature may provide a rationale for our result, as they suggest that economic agents seeking to bolster or restore their reputation may be more likely to engage in the self-promotion of falsely positive reviews \cite{luca2016, paul2021, wang2018gslda}, because, for example, a one-star increase in the average Yelp rating has been documented to be associated with 5-9\% revenue growth \cite{luca2016reviews}.
Secondly, consistent with prior observations from \cite{barbado2019framework, luca2016}, users that post more predicted fake AI-generated reviews have less established Yelp reputations as compared to those that allegedly post real content: fewer previously posted \textit{\#Reviews} (-11.37, $p$<.001), fewer \textit{\#Friends} (-5.59, $p$<.01), and fewer previously posted \textit{\#Photos} (-34.32, $p$<.001). 
Such diminished engagement levels exhibited by users that post more predicted fake AI-generated reviews might indicate an inclination toward spamming activities. It might be logical to presume that fraudsters engaging in spamming behavior would demonstrate lower levels of activity on a given platform. This might be due to their employment of rotating accounts to disseminate fabricated content, which might explain their lack of interest in cultivating reliable and trustworthy reputations within the Yelp community. At the extreme of this phenomenon, such a pattern is exacerbated by the presence of \textit{singleton-review} spammers, who generated a multitude of accounts that end up publishing a single review per account as a consequence of their fraudulent activity \cite{sandulescu2015detecting}.

%{\color{red} Therefore, we conclude that AI-generated fake reviews display similar patterns as user-generated fake reviews do at the moment. Also, these findings open new research questions about trust in purported user-generated content in the decades in which AI-generated content will proliferate.} % maybe delete this part 

%As for restaurant-based variables, we documented how {\color{red} neither} the overall average restaurant rating {\color{red} nor} the price level {\color{red} nor chain restaurants} are significantly impacted by the presence of fake AI-generated content. 
%{\color{red} The former, with a very marginal} significant mean difference of {\color{red} +.03 ($p$<.001), lacks} relevance for practical reasons since humans are not triggered by such a marginal difference. {\color{red} Instead, neither the price level nor the chain status is statistically significant. For the latter, we found a contrasting result from prior literature: \cite{luca2016} demonstrated that chain restaurants are less likely to display fake content not to erode the brand reputation. 

% REPHRASED:

Thirdly, regarding \textit{restaurant}-based variables, our study found no significant impact of fake AI-generated reviews on the overall average restaurant rating, price level, and chain status. Specifically, the very marginal difference of +.03 ($p$<.001) in \textit{AvgRating} lacks practical relevance since humans are not affected by such a small difference. 
Moreover, no difference in AI-generated fake reviews from either \textit{PriceLevel} or \textit{ChainStatus} showed any statistical significance. 
However, it is worth pointing out that our findings are in contrast with prior research concerning chain restaurants, as \cite{luca2016} found that they are less likely to display fake content to protect their brand reputation.
Next, 
%predicted AI-generated fake reviews tend to be posted in restaurants that have already displayed more reviews as compared to the restaurants in which human-generated content is prevalent (+44.71, $p$<.001). 
predicted AI-generated fake reviews tend to be associated with restaurants displaying more reviews on their Yelp web pages (\textit{\#RestReviews}, +44.71, $p$<.001).
Yet, extant literature posited that restaurants have a stronger incentive to post fake reviews when few reviews are available \cite{luca2016}, because the marginal benefit of each additional review is higher since Yelp focuses on the average rating as an indicator of customer satisfaction to be reported in restaurant web pages. 

Interestingly, by leveraging the power of the SafeGraph data, which reports the estimated per-restaurant customer visits (\textit{\#Visits}), we concluded that restaurants that displayed more AI-generated fake reviews totaled fewer customer visits (-138, $p$<.05).
To the best of our knowledge, this study represents the first analysis leveraging real users' visits to describe how fake reviews correlate with customer visits in the hospitality sector. Also, this finding raises novel research questions that aim to investigate the influence of fabricated reviews on business performance. This research direction is motivated by the need to gain a deeper understanding of the potential effects of fake reviews on consumer behavior, which can inform business strategies and policies to promote fairness and transparency in online marketplaces.

%At a descriptive level, these results lead us to conclude that Yelp is not currently being influenced by the presence of fake AI-generated content. However, constant monitoring is suggested to guarantee authentic user experiences in the future. 

Finally, we inspected the writing style of the two predicted review categories. 
As for \textit{perplexity}-based metrics, our results suggest that perplexity exhibits a downtrend pattern when applied to sequential thresholds of classification $t$. Specifically, our findings indicated that at $t$<.7, the mean \textit{Perplexity} of predicted fake AI-generated reviews was higher than for human-generated text ($p$<.01), whereas it was lower at $t$>.7 ($p$<.05). Additionally, \textit{Textual Coherence} was not found to be statistically significant at any threshold.
The pattern of \textit{Perplexity} may be explained by analyzing how large language models (LLM) are developed and generate text. 
LLM (including ChatGPT) are trained by predicting the next most likely token in a sequence of words, minimizing textual perplexity, and are more likely to output common words instead of rare words \cite{heikk2022}. 
Thus, it is reasonable to assume that AI-generated texts have lower perplexity in comparison to human-generated ones, meaning that LLM demonstrate reduced uncertainty in generating text. In other words, perplexity may reflect the likelihood of a text being machine-generated, with lower values indicating a higher probability of machine generation. In connection with this, our study reports a statistically significant downward trend in \textit{Perplexity} for AI-generated text across all thresholds $t$ (see Figure \ref{fig:writingSensitivity}). Here, higher values of $t$ can be interpreted as a higher level of confidence in classifying text as AI-generated. Therefore, we hypothesize that as our confidence in classification increases, the likelihood of misclassifying an AI-generated text decreases, thus leading to lower perplexity. Our findings are consistent with this hypothesis. In practical terms, AI-generated reviews exhibit greater grammatical correctness and predictability, yet they may lack word originality and creativity, as well as potentially be repetitive. %as a consequence of their higher certainty in generating words.

% TODO: change reference:
% https://www.technologyreview.com/2022/12/19/1065596/how-to-spot-ai-generated-text/

% READABILITY 
Next, as for \textit{readability}-based, we showed that AI-generated reviews bear a higher degree of comprehension, necessitating a lower educational grade to be understood, as measured by \textit{ARI} (-.23, $p$<.001) and \textit{\#DW} (-3.94, $p$<.001). These findings are congruent with \cite{harris2012detecting}, but different from \cite{Yoo2009}.  
Specifically, predicted human-generated reviews and AI-generated reviews score \textit{ARI} values of 7.05 and 6.82, respectively, meaning that they can be understood by average 7th and 6th-grade US students, respectively.
According to \cite{agnihotri2016online}, written content that is easily comprehensible can reduce the cognitive load on readers' information processing capabilities. As a result, such content may attract a larger readership and positively affect the perceived helpfulness of reviews. 
%Specifically, a review is considered helpful if the reader is able to comprehend the text without significant cognitive effort.
Based on this premise and on our results, we hypothesize that AI-generated fake reviews may capture readers' attention more effectively than authentic human-written reviews. Consequently, we warn that the prevalence of fake reviews may bias consumers' perceptions and intentions to visit a restaurant that has published more fake content relative to one that has not.

%%% REPHRASED
As for sentiment, aligned with \cite{Yoo2009, banerjee2014theoretical, liu2018unified}, AI-generated fake reviews presented a more positive tone (\textit{Sentiment}, +.24, $p$<.001).
Here, review spammers may be deliberately employed to alter customers' perceptions by using exaggerated language that translates into more polarized sentiment polarities \cite{paul2021}, because spammers are presumably not able to express true sentiment when writing \cite{liu2018unified}.
Based on our research findings, we conjecture that a more positive tone may be a signal of self-promoting activities. This is congruent with our previous result that AI-generated fake reviews tend to have higher average ratings (+.43, $p$<.001), as both variables are closely intertwined in the effect they measure.

%%%% PERPLEXITY:
%https://arxiv.org/abs/2301.13852-> show AI-generated fake reviews have lower perplexity. for us this is not true at the optimized threshold. however, when filtering at >.7, our results get congruent with theirs. 
% talk a little bit more 

%%%%%% OLD:
%{\color{red} Easily comprehensible text is more likely to put less cognitive load on consumers' information processing capabilities \cite{agnihotri2016online}. Easier readability may therefore attract more people to read it and increase review helpfulness. In other words, consumers would consider a review helpful only if they have been cognitively able to comprehend the text appropriately.  % rephrase. this was copy/pasted 
%Being AI-generated fake reviews simpler to read, we hypothesize that AI-generated fake reviews may quickly catch readers' attention as compared to authentic human-written reviews. This may ultimately bias people's intentions to visit a restaurant that has published more fake content with respect to another. } % REPHRASE 

% OLD
%{\color{red} Extant research argues that review spammers may deliberately be hired to alter customer perceptions using word exaggerations, implying more polarized sentiment polarities. \cite{paul2021}. In light of our findings, we hypothesize that a more positive tone may be a signal for restaurant self-promoting activities. This is also coherent with our previous finding on review ratings, which stated that fake AI-generated reviews have a higher average rating (+.43, $p$<.001), because both variables are strictly intertwined in the effect they measure.} % REPHRASE

This study is not without limitations. Firstly, we restricted the analysis to the city of New York to avoid sampling biases. However, we cannot conclude whether the results can be generalized to other areas. Secondly, we relied on the SafeGraph dataset to collect restaurant reviews. Yet, SafeGraph does not disclose the exact data collection methodology for selecting example restaurants, leaving us with uncertainty about the representativeness of the dataset.
Thirdly, we only relied on 2021 and 2022 inferential data, because GPT-3 models were proprietarily beta-released in mid-2020 \cite{brown2020}. However, years 2021 and early 2022 were still affected by the COVID-19 pandemic, thus weakening our results as compared to ordinary years, as local lockdowns may have been imposed by New York authorities, potentially changing customers' behavior. It may be reasonable to point out that results about the number of customers' visits may be subjected to changes during ordinary times. Fourthly, we highlight that when filtering out reviews at $t$=.99 only about 130 AI-generated fake reviews were singled out. This data limitation reduced statistical power in ANOVA. %Fifthly, this work relies on the assumption that elite reviews are authentic. However, this assumption cannot be validated as the necessary evidence to verify it is unavailable for researchers.
Lastly and importantly, we do not intend to provide any causal interpretation to the results found, thus limiting us in drawing 
%strong 
\textit{cause-effect} conclusions. This research should be referenced to describe patterns across the variables considered and their relationship with predicted fake AI-generated and genuine reviews. 

% FUTURE WORK: may be out of scope & we also talk about this in the conclusion
%For future work, we intend to explore whether causal patterns can be evinced from the results. In particular, the most important open question to explore will be assessing whether (and how) the presence of fake AI-generated reviews has causal relationships with the number of customers visits a restaurant receives because a visit implicitly translates into a monetary revenue conversion. 

\section{Conclusion} \label{sec:conclusion}

%While refraining from making causal claims, 
%we have delineated the accessibility of disseminating fake AI reviews generated on social media platforms. Such accessibility may lead to the proliferation of restaurant crowdturfing campaigns due to the advances in large language models, such as ChatGPT, aimed at distorting user experiences. 
Disseminating fake reviews with LLMs such as ChatGPT has become easier and cheaper than ever. Such accessibility may amplify the proliferation of "AI crowdturfing" campaigns aimed at distorting user experiences on social media.
This study proves that AI-generated fake reviews are becoming more sophisticated and can easily deceive readers. Therefore, it is imperative for policymakers to develop regulations that require online review platforms to implement tools and processes to detect and remove fake reviews. This research also underscores the need for online review platforms to invest in better detection tools for AI-generated text. As the technology used to generate fake reviews becomes more advanced, review platforms must keep pace with technological advancements to ensure they can detect and remove such content effectively. To combat this issue, we implemented AI-based detection and description of fake AI-generated content across review-based, user-based, restaurant-based, and writing-based variables, showing that fake reviews tend to have a higher rating, that users posting more AI-generated content have less established Yelp reputations, and that such AI-generated content is easier to understand as compared to the human-generated one. Notably, without providing causal claims, we also described how restaurants displaying more fake content are subjected to fewer customer visits. Up to now, no study has investigated how fake reviews correlate with customer visits. Thus, we intend to open novel research questions in this direction.

% Acknowledgments

\begin{acks}
This work was funded by Fundação para a Ciência e a Tecnologia (UID/ECO/00124/2019,
UIDB/00124/2020 and Social Sciences DataLab, PINFRA/22209/2016), POR Lisboa and
POR Norte (Social Sciences DataLab, PINFRA/22209/2016), and by Oracle via the Oracle for Research Grant providing the hardware infrastructure. 
\end{acks}

\bibliographystyle{ACM-Reference-Format}
\bibliography{sample-sigconf}


\appendix
\section{Replication Code}
Detailed replication code for the modeling part is provided at \url{https://github.com/iamalegambetti/combat-ai-restaurants.git}.
Experiments notebooks are included to quickly document model performance.

\end{document}